\documentclass[letterpaper]{article} 
\usepackage[]{aaai24}  
\usepackage{times}  
\usepackage{helvet}  
\usepackage{courier}  
\usepackage[hyphens]{url}  
\usepackage{graphicx} 
\usepackage{natbib}  
\usepackage{caption} 
\usepackage{algorithm}
\usepackage{algorithmic}
\usepackage{booktabs}
\usepackage{multirow}
\usepackage{amsmath}
\usepackage{amsfonts}
\usepackage{threeparttable}
\usepackage{makecell}
\usepackage{subfigure}
\usepackage[table]{xcolor}
\usepackage{newfloat}
\usepackage{listings}
\DeclareCaptionStyle{ruled}{labelfont=normalfont,labelsep=colon,strut=off} 
\floatstyle{ruled}
\newfloat{listing}{tb}{lst}{}
\floatname{listing}{Listing}
\title{Shrinking Your TimeStep: Towards Low-Latency Neuromorphic Object Recognition with Spiking Neural Networks}
\author{
    Yongqi Ding,
    Lin Zuo\thanks{Corresponding author: Lin Zuo.},
    Mengmeng Jing,
    Pei He,
    Yongjun Xiao
}
\affiliations{
    School of Information and Software Engineering\\ University of Electronic Science and Technology of China\\
    linzuo@uestc.edu.cn
}

\usepackage{bibentry}

\begin{document}

\maketitle

\begin{abstract}
Neuromorphic object recognition with spiking neural networks (SNNs) is the cornerstone of low-power neuromorphic computing. However, existing SNNs suffer from significant latency, utilizing 10 to 40 timesteps or more, to recognize neuromorphic objects. At low latencies, the performance of existing SNNs is significantly degraded. In this work, we propose the Shrinking SNN (SSNN) to achieve low-latency neuromorphic object recognition without reducing performance. Concretely, we alleviate the temporal redundancy in SNNs by dividing SNNs into multiple stages with progressively shrinking timesteps, which significantly reduces the inference latency. During timestep shrinkage, the temporal transformer smoothly transforms the temporal scale and preserves the information maximally. Moreover, we add multiple early classifiers to the SNN during training to mitigate the mismatch between the surrogate gradient and the true gradient, as well as the gradient vanishing/exploding, thus eliminating the performance degradation at low latency. Extensive experiments on neuromorphic datasets, CIFAR10-DVS, N-Caltech101, and DVS-Gesture have revealed that SSNN is able to improve the baseline accuracy by $6.55\%\sim21.41\%$. With only 5 average timesteps and without any data augmentation, SSNN is able to achieve an accuracy of 73.63\% on CIFAR10-DVS. This work presents a heterogeneous temporal scale SNN and provides valuable insights into the development of high-performance, low-latency SNNs.
\end{abstract}

\section{Introduction}
Brain-inspired spiking neural networks (SNNs) mimic the mammalian brain and transmit information between neurons via discrete spikes \cite{MAASS19971659}. The all-or-zero spikes and the event-driven nature make SNNs an extremely low-energy computational paradigm \cite{DENG2020294}. In addition, the temporal dynamics inherent within the spiking neurons endow SNNs with superior temporal feature extraction ability \cite{zuo2020spiking,ponghiran_spiking_2022,PALIF}. Neuromorphic data represent information in the form of 0-1 event streams similar to spikes. Computation of neuromorphic data based on SNNs holds the promise of high-performance and low-power neuromorphic computing.

\begin{figure}[t]
\centering
\includegraphics[width=0.98\columnwidth]{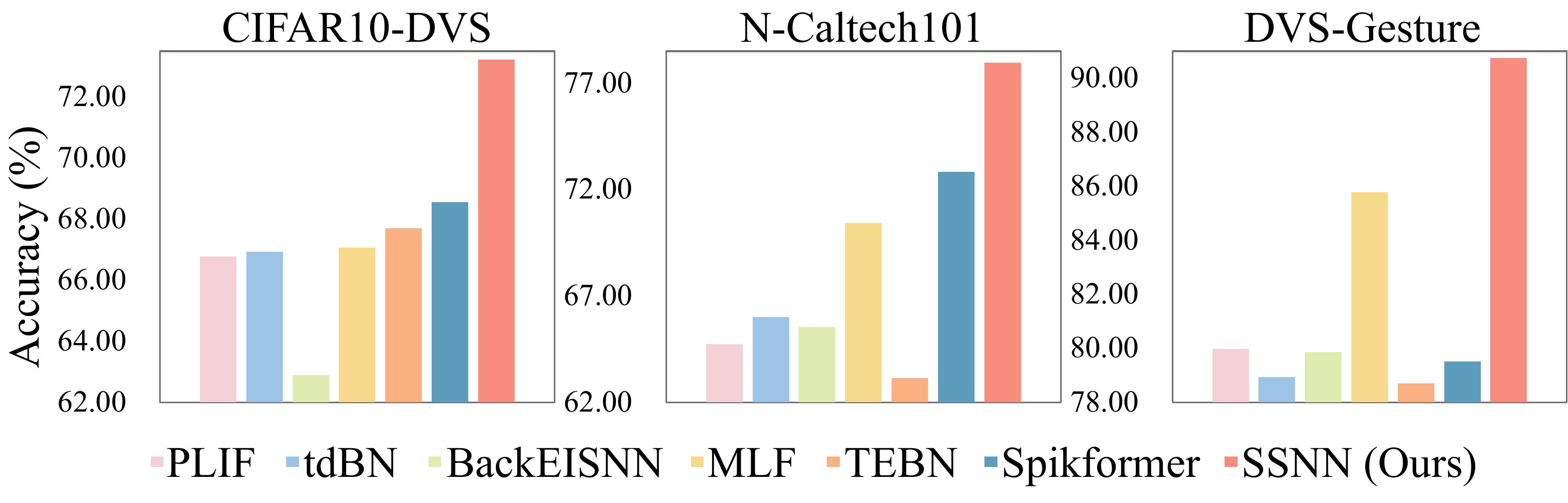}
\caption{Comparative results of the proposed SSNN and existing SNNs (with an average timestep of 5). Results show that SSNN exceeds existing SNNs by a large margin.}
\label{cal1}
\end{figure}

To improve the performance of SNNs, researchers have made efforts in various aspects such as fine-grained spiking neurons \cite{PLIF,GLIF,ding2022biologically}, modeling of biological neural properties \cite{BackEISNN,sun2023multicompartment}, and optimized training strategies \cite{TET,Guo_2022_CVPR}. Unfortunately, existing SNNs still require high latency to recognize neuromorphic objects. For instance, on the neuromorphic benchmark datasets CIFAR10-DVS \cite{CIFAR10-DVS}, N-Caltech101 \cite{N-Caltech101}, and DVS-Gesture \cite{DVS-Gesture}, 10 timesteps or more are necessary to achieve satisfactory accuracy \cite{MLF,GLIF}. With fewer timesteps, these methods are not working well, as shown in Fig. \ref{cal1}. Therefore, achieving high-performance and low-latency simultaneously needs further exploration.

Several recent works \cite{li2023unleashing,SEENN} have introduced dynamic strategies into SNNs. By employing the sample-wise timestep, the average timestep used for recognition is greatly reduced. Although this sample-wise timestep potentially limits parallel inference (batch size of 1 in \cite{li2023unleashing}), it demonstrates the potential of SNNs at very low latencies. Another work \cite{kim2022exploring} pointed out the phenomenon of temporal information concentration: during the training of SNNs, the effective information is gradually aggregated to the earlier timestep. The above work reveals the high temporal redundancy of existing SNNs, which raises a question: \textit{is it possible to reduce redundant timesteps in SNNs without sacrificing performance and parallelism?} To this end, we propose the Shrinking SNN (SSNN) for low-latency, high-performance, parallelizable neuromorphic object recognition. Specifically, inspired by the diversity of timescales in biological neurons \cite{10.3389/fncir.2020.615626}, we divide the SNN into multiple stages with progressively shrinking timesteps. To mitigate information loss due to timestep shrinkage, a simple yet effective temporal transformer is employed to smoothly transform the temporal scale of information, preserving the most valuable information with negligible additional overhead. Throughout inference, the average timestep can be significantly reduced with trivial performance degradation and without reducing parallelism.

SNNs trained from scratch with surrogate gradients suffer from two major issues: (1) mismatch between surrogate and true gradients; and (2) gradient vanishing/exploding associated with binary spikes. To further improve the performance of low-latency SNNs, we focus on these two issues by adding multiple early classifiers to SNNs to facilitate training without reducing inference efficiency. Rather than vanilla SNNs that rely only on the final global loss, losses at multiple early classifiers provide more immediate gradient feedback signals. These immediate feedback signals effectively mitigate these two major issues affecting the performance of SNNs. Meanwhile, these early classifiers also contribute to the optimization of the temporal transformer, allowing valid information to be retained more completely during timestep shrinkage.

To verify the proposed method, we conducted extensive experiments on neuromorphic datasets CIFAR10-DVS, N-Caltech101, and DVS-Gesture. At very low latencies, the performance of SSNN substantially exceeds that of existing SNNs, as shown in Fig. \ref{cal1}. In summary, the contributions of this work are as follows:
\begin{itemize}
\item We propose to reduce temporal redundancy through timestep shrinkage towards low-latency SSNN, which represents a novel SNN paradigm with heterogeneous temporal scales.
\item We add multiple early classifiers to the SNN to facilitate training, which alleviates the mismatch between the surrogate gradient and the true gradient, as well as the gradient vanishing/exploding. In this way, SSNN simultaneously achieves low latency and high performance.
\item Extensive experiments on the neuromorphic benchmark datasets CIFAR10-DVS, N-Caltech101, and DVS-Gesture validate the superior performance of our method. At an average timestep of 5, SSNN is able to improve the baseline accuracy by $6.55\%\sim21.41\%$. Compared to existing SNNs, SSNN achieves remarkable advantages at low latencies.
\end{itemize}

\section{Background}
\label{Background}

\subsection{Spiking Neuron}
Spiking neurons are in an iterative process of charging, firing, and resetting. When a spiking neuron $i$ receives input current from the presynaptic neurons, it incorporates the input current into the membrane potential for charging:
\begin{equation}
H_{i}^{l}(t) = f(U_{i}^{l}(t-1),I_{i}^{l}(t)),
\label{eq1}
\end{equation}
where $I_{i}^{l}(t)$ represents the input current, consisting of the spikes fired by the presynaptic neurons and the corresponding synaptic efficacy; $l$ and $t$ denote the layer and the timestep, respectively; $N(l-1)$ indicates the number of neurons in layer $l-1$; $H$ and $U$ denote the membrane potential of the neuron just after receiving the input current and firing spike, respectively; $f(\cdot)$ is the charging function, which varies across spiking neurons.

Once the membrane potential $H_{i}^{l}(t)$ reaches the firing threshold $\vartheta$, the spiking neuron will fire a spike and deliver it to the postsynaptic neurons:
\begin{equation}
S_{i}^{l}(t) = \Theta(H_{i}^{l}(t)-\vartheta)=\left\{
\begin{array}{cl}
1,\quad H_{i}^{l}(t) \ge \vartheta \\
0,\quad H_{i}^{l}(t) < \vartheta \\
\end{array}.
\right.
\label{eq2}
\end{equation}

After the neuron fires a spike, the membrane potential is reduced by the same magnitude as the threshold $\vartheta$, known as soft reset:
\begin{equation}
U_{i}^{l}(t) = r(H_{i}^{l}(t),S_{i}^{l}(t))=H_{i}^{l}(t)-S_{i}^{l}(t)\vartheta.
\label{eq4}
\end{equation}

In this paper, the most commonly used leaky integrate-and-fire (LIF) \cite{MLF} neuron is used as an example. The charging function of the LIF neuron is:
\begin{equation}
H_{i}^{l}(t)=\left(1-\frac{1}{\tau}\right) U_{i}^{l}(t-1)+I_{i}^{l}(t),
\label{eq5}
\end{equation}
where $\tau$ is the membrane potential time constant that controls the amount of membrane potential leakage.
\subsection{Surrogate Gradient-Based SNN Training Method}

As can be seen from Eq. (\ref{eq2}), the spike activity is not differentiable, and therefore the BP algorithm cannot work directly for SNNs. During backpropagation, the surrogate gradient-based method replaces the ill gradient of the spike activity with a smooth surrogate gradient function $h(\cdot)$. This method makes it feasible to train SNNs using the Backpropagation Through Time (BPTT) algorithm.

Specifically, the SNN calculates the spike according to Eq. (\ref{eq2}), and the derivative of the spike w.r.t. the membrane potential according to Eq. (\ref{eq6}):
\begin{equation}
\frac{\partial S_{i}^{l}(t)}{\partial H_{i}^{l}(t)} \approx \frac{\partial h(H_{i}^{l}(t), \vartheta)}{\partial H_{i}^{l}(t)},
\label{eq6}
\end{equation}
where $h(H_{i}^{l}(t),\vartheta)$ is the surrogate gradient function. In this paper, the rectangular function \cite{STBP} is used:
\begin{equation}
h(H_{i}^{l}(t),\vartheta) = \frac{1}{a} \text{sign}(|H_{i}^{l}(t)-\vartheta|<\frac{a}{2}),
\label{eq7}
\end{equation}
where $a$ serves as the hyperparameter controlling the shape of the rectangular function and is set to 1.

\begin{figure*}[t]
\centering
\includegraphics[width=0.98\textwidth]{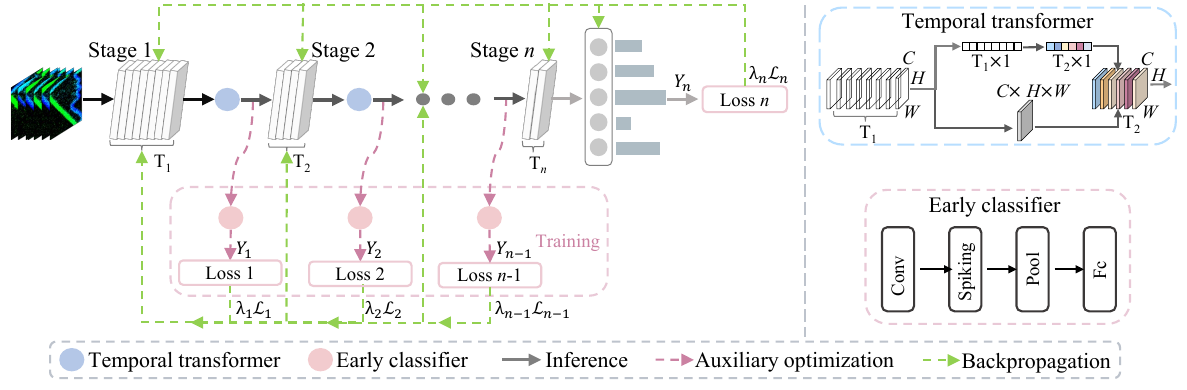} 
\caption{Overview of SSNN. The SSNN is divided into $n$ stages with gradually shrinking timesteps $\{T_1,T_2,\cdot \cdot \cdot, T_n\}$, and the temporal transformer transforms the temporal scale of the information. During training, an early classifier is added after each stage of the SSNN except the last one. The predictions generated by each early classifier are used to facilitate the optimization of the parameters by calculating the losses along with the ground truth (green arrows).}
\label{TSAC}
\end{figure*}
\section{Method}
\label{Method}
In this section, we describe the proposed SSNN for low-latency, high-performance neuromorphic object recognition. We first present two key strategies in SSNN: timestep shrinkage and early classifier, and then elaborate on the overall SSNN framework.

\subsection{Timestep Shrinkage}
As shown in Fig. \ref{TSAC}, for SNNs with arbitrary architecture and layers, we divide it into $n$ stages. 
The timestep required for each stage keeps shrinking as the network layers get deeper, that is, $T_1>T_2>\cdot \cdot \cdot>T_n$. This practice follows the basic principle that the input part of the SNN requires a large timestep to extract sufficiently valuable features from the input, while the remaining part of the SNN uses fewer timesteps to reduce the inference latency. At the same time, the gradual shrinkage of the timestep, rather than a sudden and drastic reduction, allows the information transmitted to be effectively retained when the temporal scale is reduced, thus preventing performance degradation.

\subsubsection{Temporal Transformer}
For stages with different timesteps, the temporal dimension of the data processed is different. For example, assume that the timesteps of the two stages are $T_1$ and $T_2$, and the data that can be processed are $\boldsymbol{I}_1 \in \mathbb{R}^{T_1 \times C \times H \times W}$ and $\boldsymbol{I}_2 \in \mathbb{R}^{T_2 \times C \times H \times W}$, where $C$, $H$, and $W$ denote the channel, height, and width of the intermediate feature map, respectively. To allow smooth transmission of information between two stages with different temporal scales, the dimensionality of the output of the pre-stage must be converted to an acceptable dimension for the post-stage and then transmitted to the post-stage. Therefore, how to transform information between different temporal scales and maximize the preservation of valid information is a non-trivial problem. To enable the smooth transformation of information across temporal scales, we propose a lightweight and tractable yet effective temporal transformer. Denoting the output of the pre-stage as $\boldsymbol{O}_1 \in \mathbb{R}^{T_1 \times C \times H \times W}$, the temporal transformer is described as follows, and the graphical representation is shown at the top right of Fig. \ref{TSAC}.

First, the timestep-wise global average descriptor of $\boldsymbol{O}_1$ is calculated within $T_1$ timesteps:
\begin{equation}
\boldsymbol{O}_1^{avg} = \frac {1}{C \times H \times W} \displaystyle \sum_{i=1}^{C} \sum_{j=1}^{H} \sum_{k=1}^{W}\boldsymbol{O}_{1,i,j,k},
\label{eq8}
\end{equation}
where $\boldsymbol{O}_1^{avg} \in \mathbb{R}^{T_1 \times 1}$ is the timestep-wise global average descriptor. Then, the temporal score $\boldsymbol{d} \in \mathbb{R}^{T_2 \times 1}$ for $T_2$ timesteps can be obtained using a nonlinear transformation:
\begin{equation}
\boldsymbol{d} = \text{softmax} (\boldsymbol{W}\boldsymbol{O}_1^{avg}),
\label{eq9}
\end{equation}
where $\boldsymbol{W} \in \mathbb{R}^{T_2 \times T_1}$ is the learnable weight of the nonlinear transformation. The softmax function keeps the sum of temporal scores to 1 to ensure the complete assignment of information. The temporal score $\boldsymbol{d}$ serves as the basis for reassigning $\boldsymbol{O}_1$ in $T_2$ timesteps. Calculate the sum of $\boldsymbol{O}_1$ over $T_1$ timesteps as the total information and 
assign it to $T_2$ timesteps based on $\boldsymbol{d}$ to obtain $\boldsymbol{I}_2 \in \mathbb{R}^{T_2 \times C \times H \times W}$:
\begin{equation}
\boldsymbol{I}_{2,t} = \boldsymbol{O}_1^{total} \odot \boldsymbol{d}_t  = \sum_{t^{'}=1}^{T_1}\boldsymbol{O}_{1,t^{'}} \odot \boldsymbol{d}_t,
\label{eq10}
\end{equation}
where $\boldsymbol{O}_1^{total} \in \mathbb{R}^{C \times H \times W}$ is the total information; $\odot$ denotes the multiplication with the broadcast mechanism. 

The $\boldsymbol{I}_2$ obtained after the above transformation has a post-stage compatible temporal dimension and can be used as input to the post-stage to continue the forward inference. In this way, the timestep can be continuously shrunk to an ultra-low value without causing significant information loss during the shrinking process.

\subsubsection{Average Timestep and Overhead Analysis}
Assuming that the SNN is divided into $n$ stages, each stage containing $n_i$ computational units (a convolutional layer and a layer of spiking neurons), requiring timesteps of $\{T_1,T_2,\cdot\cdot\cdot,T_n\}$, then the average timestep for inference using the SNN with timestep shrinkage can be approximated as:
\begin{equation}
T_{avg}=\frac {\sum_i^n {n_iT_i}} {\sum_i^n n_i}.
\label{eq12}
\end{equation}
Note that the fully connected layer used for classification (which runs at the minimum timestep $T_n$) is not considered in Eq. (\ref{eq12}), so the actual average timestep is less than $T_{avg}$. For an SNN that does not use timestep shrinkage, the average timestep is equal to the timestep of each stage. In subsequent experiments, the SNNs with and without timestep shrinkage are compared at the same average timestep to confirm the effectiveness of timestep shrinkage.

The number of learnable parameters introduced by the temporal transformer is positively correlated with the timesteps of the two adjacent stages. Within the whole SNN, the number of additional introduced parameters $w_{add}=\sum_1^{n-1} {T_iT_{i+1}}$. For neuromorphic data, the timestep before and after the shrinkage is typically less than or equal to 10. The number of stages $n$ that divide an SNN is generally kept between 3 and 5. Accordingly, $w_{add}$ is kept to the order of a hundred, which is negligible compared to the millions of total parameters in the SNN.

\subsection{Early Classifier}

SNNs trained with the surrogate gradient bypass the non-differentiability of the spike activity, but the mismatch between the surrogate gradient and the true gradient limits the performance of the SNN. On the other hand, binary spikes make SNNs suffer from more severe gradient vanishing/exploding than artificial neural networks \cite{SEWResNet}. To improve the performance of SSNN and eliminate the performance degradation at low latency, we mitigate these two major problems by adding multiple early classifiers during training. These losses at the early classifiers deliver more immediate gradient signals, effectively alleviating both problems while contributing to the optimization of the temporal transformer.

During training, the output of each stage, except the last one, is passed to an early classifier for auxiliary optimization (rose arrows in Fig. \ref{TSAC}). Each early classifier consists of a convolutional layer, a spiking neuron layer, and a fully connected layer, similar to \cite{BranchyNet}, capable of predicting the object based on intermediate features generated by the corresponding stage (the bottom right of Fig. \ref{TSAC}). In this paper, early classifiers have the same structure and can actually be customized depending on the difficulty of the task and the location of the early classifiers. Alternatively, a globally shared early classifier can be used after each stage to reduce the overhead during training. We use the same structure of early classifiers rather than a deliberate setting in order to facilitate a clearer focus on the effect of this strategy rather than the structure of each early classifier. The output of the last stage predicts the object directly through the fully connected layer and without passing it to the early classifier. Thus, each stage will produce a prediction specific to the input and calculate the loss with the ground truth. 

The output is decoded in a rate-based manner, that is, the average of multiple timestep outputs is used to calculate the loss along with the ground truth. The total loss is expressed as the weighted sum of the losses at each stage:
\begin{equation}
\mathcal{L}_{total}=\sum_i^n {\lambda_i \mathcal{L}_i(\frac{1}{T_i}\sum_t^{T_i} {\boldsymbol{Y}_{i,t},\hat{\boldsymbol{Y}}})},
\sum_i^n \lambda_i = 1,
\label{eq14}
\end{equation}
where $\boldsymbol{Y}_{i,t}$ is the output of the $i$-th stage through early classifier at timestep $t$; $\hat{\boldsymbol{Y}}$ is the ground truth; $\mathcal{L}_i$ denotes the loss of the $i$-th stage; $\lambda_i$ is the coefficient corresponding to the loss. In this paper, all losses are set to the cross-entropy (CE) loss. It is worth noting that this strategy is compatible with specially designed loss functions such as TET \cite{TET} and IM-Loss \cite{NEURIPS2022_010c5ba0}. The total loss is backpropagated to each layer in the SNN based on the surrogate gradient function and the BPTT algorithm. 
The losses of each stage act on the optimization of the parameters in all its preceding layers. Compared with methods that rely only on the final output, these immediate gradients mitigates the impact of gradient mismatch accumulation and gradient vanishing/exploding.

The output of the first stage is based only on the extracted primary features, and the corresponding loss can only facilitate the updating of local parameters. The later stages produce output on the basis of high-level features, and their losses provide broader perspectives (green arrows in Fig. \ref{TSAC}). In the subsequent experiments, we experimentally investigate the effect of different loss coefficients $\lambda$ on the performance and show that this strategy is not sensitive to $\lambda$ with great generalization.

\subsubsection{Additional Overhead Analysis}
The early classifiers added after each stage are only used for auxiliary optimization and do not affect the inference process. Therefore, this strategy do not incur additional computational overhead during inference.
\begin{algorithm}[tb]
\caption{Training framework for SSNN} 
\label{alg1}
\begin{algorithmic}[1]
\REQUIRE input $\boldsymbol{X}$, label $\hat{\boldsymbol{Y}}$, the number of stages $n$ of SNN, the timestep of each stage $\{T_1,T_2,\cdot \cdot \cdot, T_n\}$.
\ENSURE Update network parameters.
\STATE Initialize network parameters $\boldsymbol{W}$;
\STATE $\boldsymbol{I}_1=\boldsymbol{X}$;
\FOR {$i=1$ to $n-1$} 
\STATE $\boldsymbol{O}_i \leftarrow stage_i(\boldsymbol{I}_i)$; // Calculate stage output
\STATE $\boldsymbol{I}_{i+1} \leftarrow$ Eq. (\ref{eq8}-\ref{eq10}); // Transform output and shrink timestep
\STATE $\boldsymbol{Y}_{i} \leftarrow EC_i(\boldsymbol{I}_{i+1})$; // Calculate the output of early classifier
\STATE $\boldsymbol{\mathcal{L}_{i}} \leftarrow \boldsymbol{\mathcal{L}}_i(\frac{1}{T_{i+1}}\sum_t^{T_{i+1}} \boldsymbol{Y}_{i,t},\hat{\boldsymbol{Y}})$; // Calculate the loss of early classifiers
\ENDFOR
\STATE $\boldsymbol{O}_n \leftarrow stage_n(\boldsymbol{I}_{n})$;
\STATE $\boldsymbol{Y}_n \leftarrow fc(\boldsymbol{O}_n)$;
\STATE $\boldsymbol{\mathcal{L}_{n}} \leftarrow \boldsymbol{\mathcal{L}}_n(\frac{1}{T_n}\sum_t^{T_n} \boldsymbol{Y}_{n,t},\hat{\boldsymbol{Y}})$;
\STATE $\boldsymbol{\mathcal{L}_{total}} \leftarrow$ Eq. (\ref{eq14});
\STATE Update parameters $\boldsymbol{W}$ based on the BPTT algorithm.
\end{algorithmic}
\end{algorithm}
\subsection{The SSNN Framework}

SSNN achieves low-latency inference through timestep shrinkage and high performance with multiple early classifiers, as shown in Fig. \ref{TSAC}. The temporal transformer transforms the output of the pre-stage to make it compatible with the temporal scale of the post-stage. With the immediate gradient of the early classifiers, it can facilitate the optimization of the learnable parameters in the temporal transformer, thus favoring the preservation of information during temporal scale transformation. Specifically, the output of each stage is transformed by the temporal transformer first, and then the transformed information is fed to both the early classifier and the post-stage. In this way, the input received by each early classifier has been shrunk in timestep. The training framework for SSNN is presented in Algorithm \ref{alg1}.

\section{Experiments}
\label{Experiments}
Experiments were conducted on CIFAR10-DVS \cite{CIFAR10-DVS}, N-Caltech101 \cite{N-Caltech101}, and DVS-Gesture \cite{DVS-Gesture}. The preprocessing of neuromorphic data is similar to MLF \cite{MLF}, integrating the event streams into frames and then downsampling without any data augmentation. All experiments were repeated three times with different random seeds to reduce randomness. Two architectures, VGG-9 and ResNet-18, are used in the experiments to demonstrate the generalizability of the proposed methods. Both of these network structures are evenly divided into four stages. The detailed data processing process and implementation details can be found in \textbf{Supplementary Material A}.
\begin{figure}[!tb]
\centering
\includegraphics[width=0.98\columnwidth]{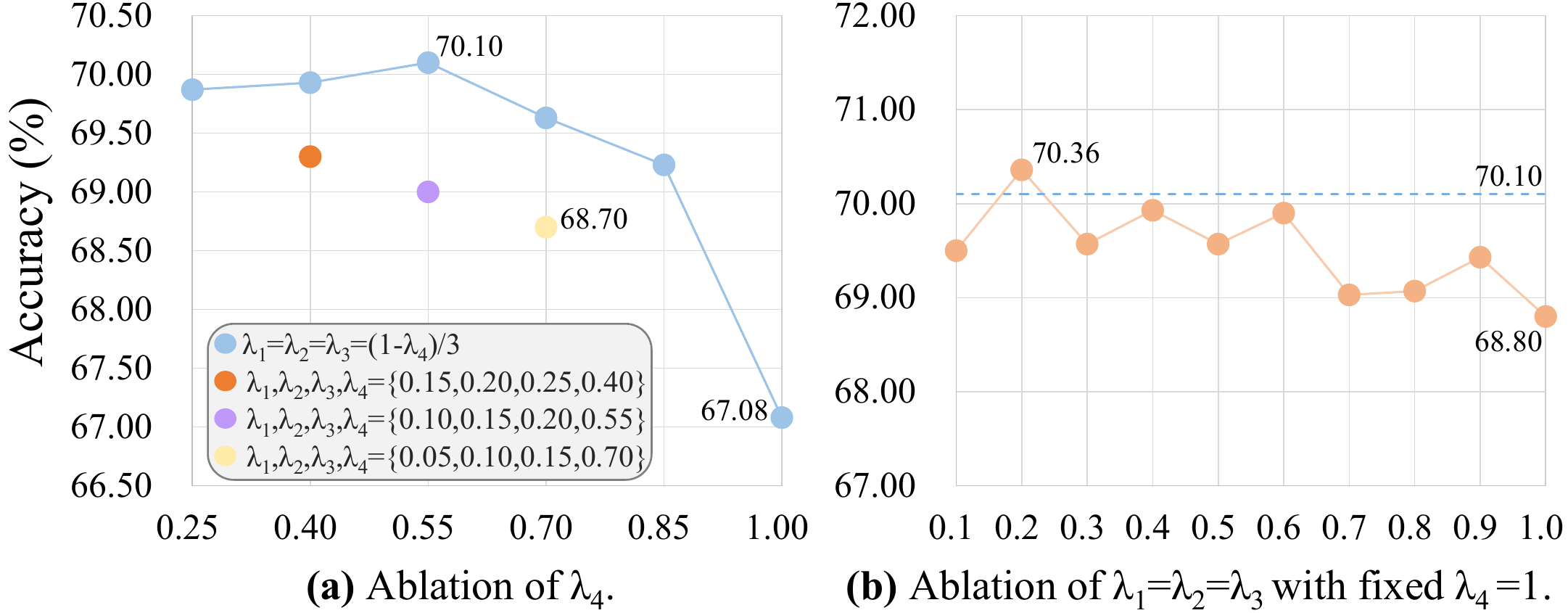}
\caption{Influence of $\lambda$ on performance. The accuracy remains at [68.70\%,70.36\%] as long as $\lambda_1$, $\lambda_2$ and $\lambda_3$ are not zero, indicating that our method is not sensitive to $\lambda$.}
\label{lambda}
\end{figure}
\subsection{Ablation Studies}

\textbf{Influence of $\lambda$.} We first explore the influence of $\lambda$ under two settings: (a) $\lambda_1=\lambda_2=\lambda_3=(1-\lambda_4)/3$; (b) The coefficients of the losses gradually increase as the layers get deeper, i.e., $\lambda_1<\lambda_2<\lambda_3<\lambda_4$. The experiments were performed on CIFAR10-DVS using VGG-9 with a timestep of 5. The experimental results are shown in Fig. \ref{lambda}(a). When $\lambda_4=1, \lambda_1=\lambda_2=\lambda_3=0$, only the final loss works (auxiliary optimization does not work) with an accuracy of 67.08\%. As $\lambda_1$, $\lambda_2$ and $\lambda_3$ increases, the auxiliary losses gradually contribute to the optimization of the parameters and the performance improves. When $\lambda_4\in[0.25,0.70]$, the accuracy fluctuates between 68.70\% and 70.10\%, which indicates that our method is not sensitive to $\lambda$ (as long as $\lambda_1$, $\lambda_2$ and $\lambda_3$ are not 0) and has great generalizability. In addition, we step over the constraint that the weighted sum is 1 and fix $\lambda_4=1$ to make $\lambda_1=\lambda_2=\lambda_3$ to further explore the influence of $\lambda$. As shown in Fig. \ref{lambda}(b), the accuracy is kept between 68.80\% and 70.36\%, which is quite stable. In subsequent experiments, we fixed $\lambda_1=\lambda_2=\lambda_3=0.15,\lambda_4=0.55$ for stable results.

\begin{table}[t]
\tabcolsep=0.004\columnwidth
 \centering
 \begin{tabular}{ccccc}
  \toprule
 Method & CIFAR10-DVS & N-Caltech101 & DVS-Gesture\\
  \midrule
  VGG-9 & 67.08 & 66.45 & 80.29\\
  +TS & $71.53_{+4.45}$ & $75.42_{+8.97}$ & $85.42_{+5.13}$\\
  +EC & $70.10_{+3.02}$ & $70.13_{+3.68}$ & $84.49_{+4.20}$\\
  SSNN & $\textbf{73.63}_{+6.55}$ & $\textbf{77.97}_{+11.52}$ & $\textbf{90.74}_{+10.45}$\\
  \hline
  ResNet-18 & 58.60 & 52.81 & 65.17\\
  +TS & $61.97_{+3.37}$ & $62.62_{+9.81}$  & $72.68_{+7.51}$\\
  +EC & $64.30_{+5.70}$ & $57.29_{+4.48}$ & $77.43_{+12.26}$\\
  SSNN & $\textbf{68.10}_{+9.50}$ & $\textbf{67.51}_{+14.70}$ & $\textbf{86.58}_{+21.41}$\\
  \bottomrule
 \end{tabular}
 \caption{Average test accuracy (\%) for ablation studies of timestep shrinkage (TS), early classifier (EC), and SSNN.}
 \label{ablation}
\end{table}
\begin{table*}[!tb]
 \centering
 \begin{threeparttable}
 \begin{tabular}{cccccc}
  \toprule
 Dataset & Method & Network & Average timestep & Accuray (\%)\\
  \midrule
  \multirow{14}{*}{CIFAR10-DVS} 
  &DSR \cite{DSR} & VGG-11\tnote{$\dag$} & 20 & 77.27 \\
  &GLIF \cite{GLIF} & 7B-wideNet & 16 & 76.80 \\
  &TET \cite{TET} & VGGSNN & 10 & 77.33 \\
  &SLTT \cite{SLTT} & VGG-11 & 10 & 77.17 \\
  &SLSSNN \cite{SLSSNN} & VGG-16 & 8 & 77.50 \\
  &AutoSNN \cite{AutoSNN} & AutoSNN\tnote{$\dag$} & 8 & 72.50 \\
  \cline{2-5}
  & STBP-tdBN \cite{tdBN_2021} & VGG-9\tnote{*} & 5 & 66.93 \\
  & PLIF \cite{PLIF} & VGG-9\tnote{*} & 5 & 66.77 \\
  & BackEISNN \cite{BackEISNN} & VGG-9\tnote{*} & 5 & 62.90 \\
  & MLF \cite{MLF} & VGG-9\tnote{*} & 5 & 67.07 \\
  & TEBN \cite{TEBN} & VGG-9\tnote{*} & 5 & 67.70 \\
  &Spikformer \cite{Spikformer} & Spikformer\tnote{*$\dag$} & 5 & 68.55 \\
  \cline{2-5}
  &\multirow{2}{*}{\textbf{SSNN} (Ours)}  & VGG-9 & 5 & \textbf{73.63} \\&& VGG-9 & 8 & \textbf{78.57}\\
  \hline
  \multirow{10}{*}{N-Caltech101}
  &EventMix \cite{eventmix} & ResNet18\tnote{$\dag$} & 10 & 79.47 \\
  &tdBN+NDA \cite{NDA} & VGG-11\tnote{$\dag$} & 10 & 78.20 \\
  \cline{2-5}
  &STBP-tdBN \cite{tdBN_2021}  & VGG-9\tnote{*} & 5 & 66.01 \\
  &PLIF \cite{PLIF}  & VGG-9\tnote{*} & 5 & 64.73 \\
  &BackEISNN \cite{BackEISNN} & VGG-9\tnote{*} & 5 & 65.53 \\
  &MLF \cite{MLF} & VGG-9\tnote{*} & 5 & 70.42 \\
  &TEBN \cite{TEBN} & VGG-9\tnote{*} & 5 & 63.13 \\
  &Spikformer \cite{Spikformer} & Spikformer\tnote{*} & 5 & 72.83 \\
  \cline{2-5}
  &\multirow{2}{*}{\textbf{SSNN} (Ours)}  & VGG-9 & 5 & \textbf{77.97} \\&& VGG-9 & 8 & \textbf{79.25}\\
  \hline
  \multirow{9}{*}{DVS-Gesture}
  &teacher default-KD \cite{KD} & ResNet18 & 16 & 96.88 \\
  \cline{2-5}
  & STBP-tdBN \cite{tdBN_2021} & VGG-9\tnote{*} & 5 & 78.94 \\
  & PLIF \cite{PLIF} & VGG-9\tnote{*} & 5 & 79.98 \\
  & BackEISNN \cite{BackEISNN} & VGG-9\tnote{*} & 5 & 79.86 \\
  & MLF \cite{MLF} & VGG-9\tnote{*} & 5 & 85.77 \\
  & TEBN \cite{TEBN} & VGG-9\tnote{*} & 5 & 78.70 \\
  &Spikformer \cite{Spikformer} & Spikformer\tnote{*} & 5 & 79.52 \\
  \cline{2-5}
  &\multirow{2}{*}{\textbf{SSNN} (Ours)}  & VGG-9 & 5 & \textbf{90.74} \\&& VGG-9 & 8 & \textbf{94.91}\\
  \bottomrule
 \end{tabular}
\end{threeparttable}
\caption{Comparative results with existing methods. * denotes self-implementation results. $\dag$ indicates using data augmentation.}
 \label{comparative }
\end{table*}
\textbf{Comparison with the baseline SNN.} We compare the proposed method with the baseline SNN on three datasets. The timesteps of the four stages of VGG-9 and ResNet-18 are set to $\{8,6,4,2\}$ and $\{8,6,4,1\}$, and their average timesteps are 5 and 4.94, respectively. The baseline timestep is set to 5. As shown in Table \ref{ablation}, the baseline does not perform well on the three datasets, indicating that the vanilla training method cannot achieve satisfactory performance at low latency. By separately employing timestep shrinkage and early classifier, there is a noticeable improvement in the accuracy, demonstrating the effectiveness of these methods in enhancing the performance of low-latency SNNs. The optimal performance is obtained by SSNN, which considerably outperforms the baseline. When inference is performed on DVS-Gesture with ResNet18, the accuracy of SSNN even exceeds that of the baseline by 21.41\%. To intuitively illustrate the superiority of the proposed method, we provide the accuracy change curves during training in \textbf{Supplementary Material B}.

\begin{figure}[!tb]
\centering
\includegraphics[width=0.98\columnwidth]{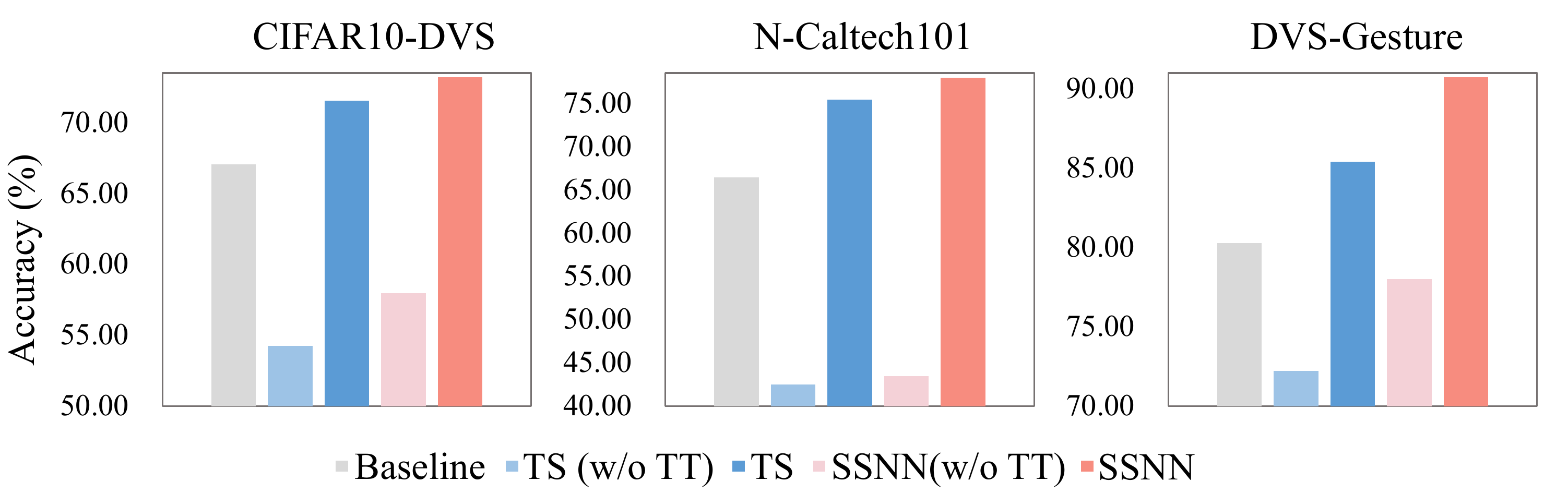}
\caption{Influence of the temporal transformer (TT). The performance of SNNs is severely degraded and inferior to the baseline without the temporal transformer.}
\label{alm}
\end{figure}

\textbf{Effectiveness of the temporal transformer.} To evaluate the role of the temporal transformer, we compare the performance of timestep shrinkage and SSNN with and without the temporal transformer. Assuming $T_2$ is the shrunken timestep, the information of the first $T_2$ timesteps is taken directly for subsequent processing in the absence of the temporal transformer. VGG-9 was used for the experiments and the average timestep was set to 5. The experimental results are shown in Fig. \ref{alm}. In the absence of the temporal transformer, shrinking the timestep leads to degraded performance, and even the accuracy of SSNN is lower than the baseline. This illustrates that information must be rationally reassigned when the temporal scale has changed, and that direct truncation of information will lead to dramatic loss of information. Considerable performance gains are achieved by employing the temporal transformer, which demonstrates the crucial role of the temporal transformer.

\begin{figure}[!tb]
\centering
\includegraphics[width=0.8\columnwidth]{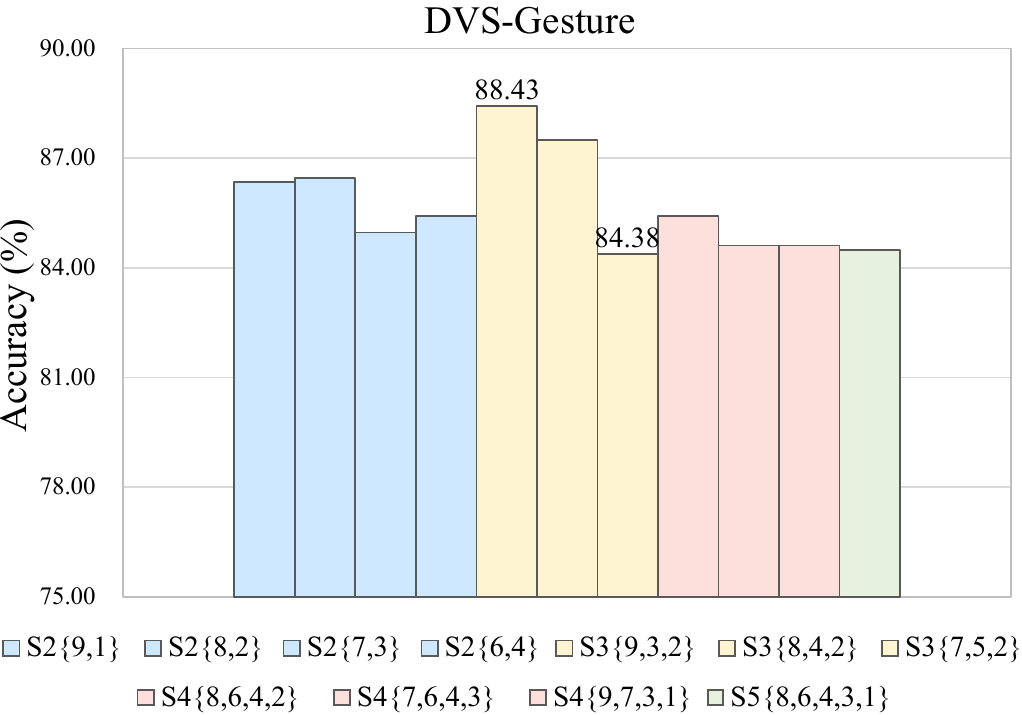}
\caption{Influence of stage division and stage-wise timesteps, S$n\{t_1,t_2,... ,t_n\}$ denotes division into $n$ stages and stage-wise timestep of $\{t_1,t_2,... ,t_n\}$. Results indicate that our method is not sensitive to different stage divisions and stage-wise timesteps.}
\label{stage}
\end{figure}
\textbf{Influence of stage division and stage-wise timesteps.} We divided VGG-9 into 2, 3, 4, and 5 stages and set different stage-wise timesteps by keeping the average timestep at 5 to explore the influence of these factors on the timestep shrinkage strategy. The 8 convolution layers in VGG-9 are evenly divided into 2 and 4 stages. In case of division into 3 or 5 stages, each stage contains the number of convolution layers as \{3,3,2\} or \{2,2,2,1,1\}. In total, we explored a total of 11 different settings, see Fig. \ref{stage}. As shown in Fig. \ref{stage}, the accuracy is in the range of 84.38\% and 88.43\%, without drastic changes as the way of division and stage-wise timesteps vary. In both settings, the accuracy far exceeds the baseline accuracy of 80.29\%. This verifies that our method is not sensitive to these settings and has great generalizability. It is worth noting that when divided into two stages with timesteps of 9 and 1, it achieved a surprising accuracy of 86.34\%. This illustrates that the timestep shrinkage strategy can work with flexible timestep settings, even for substantial temporal scale shrinkage.

\subsection{Comparison with Existing Methods}
Since existing methods typically work with larger timesteps, we reproduced some of the methods for a fair comparison, and the details can be found in \textbf{Supplementary Material C}. The comparative results are shown in Table \ref{comparative }.

\subsubsection{CIFAR10-DVS} As shown in Table \ref{comparative }, SSNN outperforms all other methods at an average timestep of 5. Even though Spikformer \cite{Spikformer} uses a more complex transformer structure and data augmentation, the accuracy of our SSNN is still 5.08\% higher than that of Spikformer. When the average timestep is increased to 8, SSNN achieves an accuracy of 78.57\%, surpassing SLTT \cite{SLTT}, TET \cite{TET}, GLIF \cite{GLIF}, and DSR \cite{DSR} at higher latencies.

\subsubsection{N-Caltech101} With an average timestep of 5, our SSNN achieves an accuracy of 77.97\%, which is 5.32\% higher than that of Spikformer \cite{Spikformer}. EventMix \cite{eventmix} and NDA \cite{NDA} achieved accuracies of 79.47\% and 78.20\% at timestep 10, respectively, using specially designed data augmentation methods. Our SSNN achieves an accuracy of 79.25\% at an average timestep of 8 without data augmentation, which exceeds NDA \cite{NDA} at a timestep of 10 and is only 0.22\% worse than EventMix \cite{eventmix}.

\subsubsection{DVS-Gesture} When the timestep is 5, only MLF \cite{MLF} can achieve 85.77\% accuracy, while the other methods have poor recognition performance. At this point, our SSNN was able to achieve an accuracy of 90.74\%, exceeding MLF \cite{MLF} by 4.97\%. For an average timestep of 8, SSNN is able to achieve an accuracy of 94.91\%, which is only 1.97\% lower than that of teacher default-KD \cite{KD} at a timestep of 16, while the required inference latency is reduced by half. In \textbf{Supplementary Material E}, we further compare the performance of our method with that of Spikformer at 16 time steps and confirm the mutual enhancement effect of the proposed method with Spikformer.
\begin{figure}[!t]
\centering
\includegraphics[width=0.98\columnwidth]{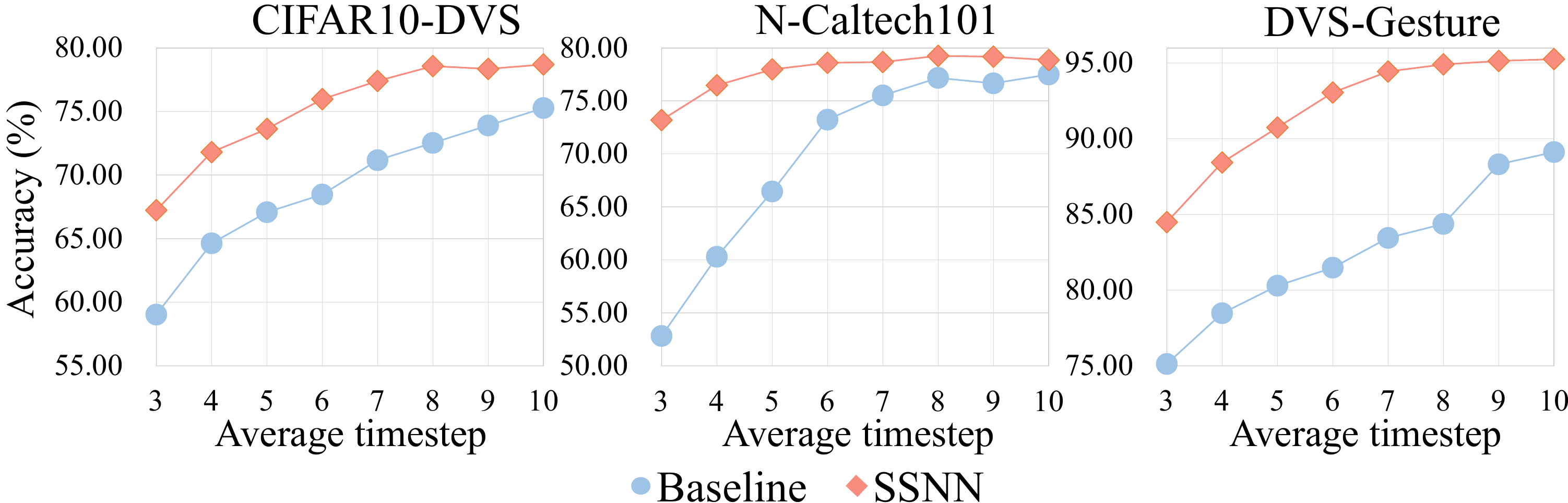}
\caption{Influence of average timestep on performance. SSNN consistently outperforms the baseline and shows significant advantages, especially at low latencies.}
\label{timestep}
\end{figure}
\subsection{Influence of Average Timestep}
Here, we evaluate the influence of average timestep on the performance of SSNN. The experiments were conducted using the VGG-9 with controlled average timesteps ranging from 3 to 10, and the detailed stage-wise timestep settings can be found in \textbf{Supplementary Material D}. The experimental results are shown in Fig. \ref{timestep}. When the average timestep is slightly larger, SSNN exceeds the baseline by a small margin; as the average timestep gradually decreases, the performance gap between the two grows more significant. It is worth noting that the performance of SSNN at low latency (e.g., 4/5/6) exceeds that of the baseline at high latency (e.g., 7/8/10, which is the timestep of the first stage when the average timestep is 4/5/6), revealing that our SSNN effectively avoids the performance degradation associated with timestep shrinkage.

\subsection{Visualization}
\label{Visualization}
\begin{figure}[!t]
\centering
\includegraphics[width=0.95\columnwidth]{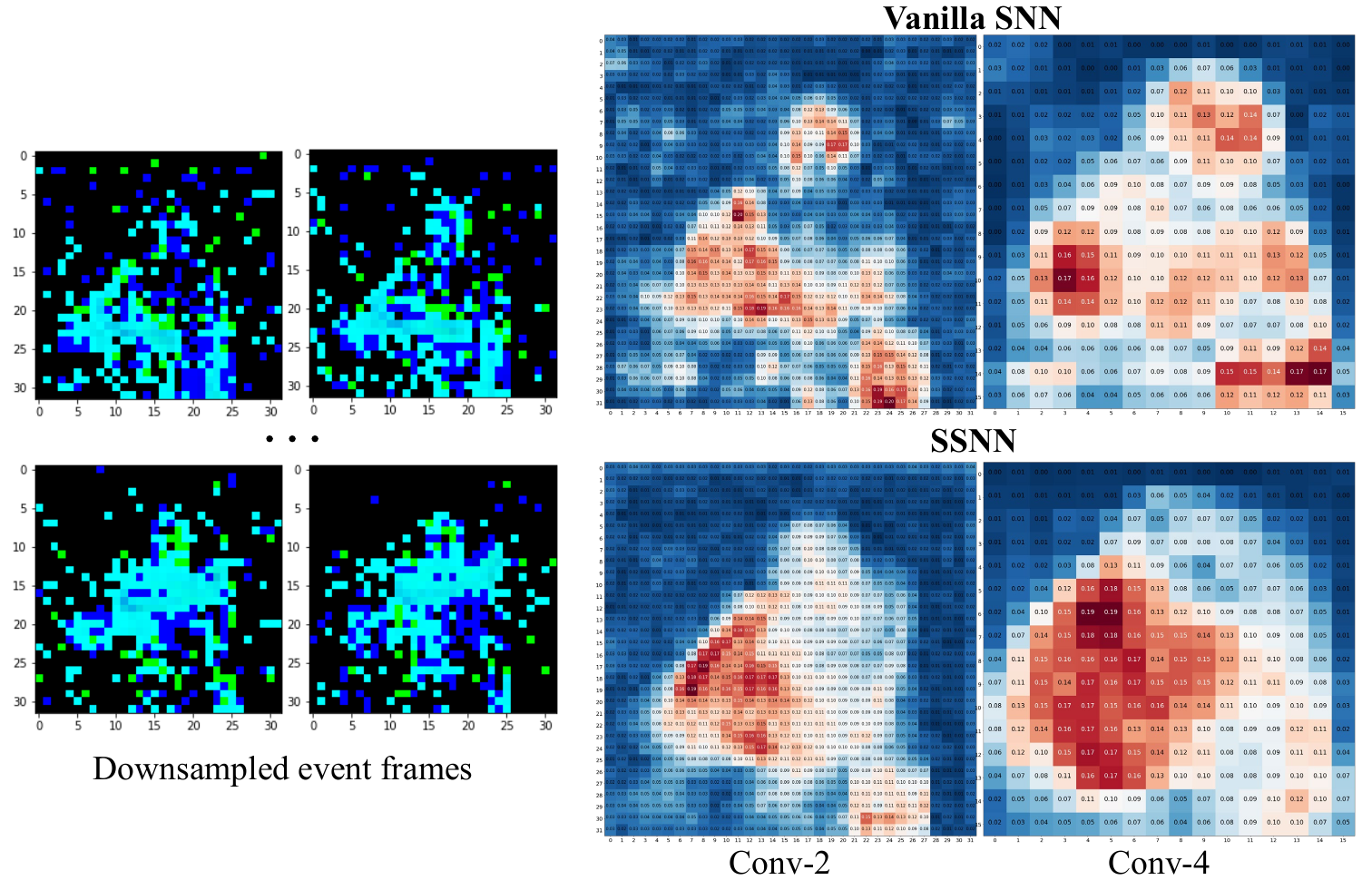}
\caption{Visualization of spike firing rate on DVS-Gesture. Compared to the vanilla SNN, SSNN can accurately focus on the hand region, which is crucial for gesture recognition.}
\label{firing_rate}
\end{figure}

To understand more clearly the feature extraction capability of SSNN, similar to \cite{10032591}, we visualize the spike firing rates of the second and fourth convolutional layers in both vanilla SNN and SSNN. As shown in Fig. \ref{firing_rate}, the vanilla SNN cannot focus on the waving hand and the focus area is scattered. In contrast, SSNN can exactly capture the hand region (which is critically important for gesture recognition) and therefore can extract features more accurately and perform better for recognition.

\section{Conclusion}
\label{Conclusion}
In this work, we propose SSNN for low-latency, high-performance neuromorphic object recognition. On the one hand, SSNN reduces temporal redundancy by progressively shrinking the timestep to achieve low-latency inference. To alleviate the information loss during timestep shrinkage, a simple yet effective temporal transformer is employed to smoothly transform the temporal scale of the information. On the other hand, SSNN mitigates the gradient mismatch and gradient vanishing/exploding by using multiple early classifiers during training to improve the recognition performance under low latency. Extensive experiments have confirmed the effectiveness of the proposed SSNN. At very low latency, the performance of SSNN far exceeds that of existing SNNs. We expect that our work can contribute to the study of heterogeneous temporal scale SNNs and inspire the development of ultra-low latency, high-performance SNNs.
\section{Acknowledgments}
This work was supported by the National Natural Science Foundation of China (Grant No. 62276054, 61877009), and the Sichuan Science and Technology Program (Grant No. 2023YFG0156).

\appendix
\section{Supplementary Material}
\label{Supplementary Material}
\subsection{A Details of Experiments}
\subsection{A.1 Dataset}
We evaluate the proposed method on three neuromorphic benchmark datasets, CIFAR10-DVS, N-Caltech101, and DVS-Gesture.

\textbf{CIFAR10-DVS:} CIFAR10-DVS \cite{CIFAR10-DVS} is a neuromorphic version of the CIFAR10 dataset and is a benchmark for evaluating the performance of SNNs. There are 10,000 samples in CIFAR10-DVS, and each sample is represented as an event stream. The spatial size of all samples is $128 \times 128$, and there are 10 classes of objects. Each event stream $x \in [t,x,y,p]$ indicates the change in pixel value or brightness at location $[x,y]$ at the moment $t$ relative to the previous moment. $p$ represents polarity, and positive polarity indicates an increase in pixel value or brightness, and vice versa.

The data in CIFAR10-DVS are preprocessed as in \cite{MLF}. The original event stream is split into multiple slices in 10ms increments, and each slice is integrated into a frame and downsampled to $42 \times 42$. CIFAR10-DVS is divided into training and test sets in the ratio of 9:1.

\textbf{N-Caltech101:} N-Caltech101 \cite{N-Caltech101} is a neuromorphic version of Caltech101 with 101 categories and 8709 samples with a spatial size of $180 \times 240$. The pre-processing of the N-Caltech101 data is the same as for CIFAR10-DVS. The event stream is integrated into one frame every 10ms using the SpikingJelly package \cite{SpikingJelly} and downsampled to $42 \times 42$. The ratio of 9:1 is used to divide it into a training set and a test set.

\textbf{DVS-Gesture:} DVS-Gesture \cite{DVS-Gesture} is a neuromorphic dataset for gesture recognition. DVS-Gesture contains a total of 11 event stream samples of gestures, 1176 for training and 288 for testing, with a spatial size of $128 \times 128$ for each sample. For DVS-Gesture data, the event stream is integrated into frames in 30ms units and downsampled to $32 \times 32$.

\subsection{A.2 Implementation Details}
The experiments were conducted with the PyTorch package. All models were run on NVIDIA TITAN RTX with 100 epochs of training. The initial learning rate is set to 0.1 and decays to one-tenth of the previous rate every 30 epochs. The batch size is set to 64. The stochastic gradient descent optimizer was used with momentum set to 0.9 and weight decay to 1e-3. For LIF neurons, set the membrane potential time constant $\tau = 2.0$ and the threshold $\vartheta = 1.0$.  

\subsection{A.3 Network Structures}
In our experiments, we use both VGG-9 and ResNet-18 architectures to validate our method. The specific structure of these two networks and the way the stages are divided are shown in Table \ref{model}. For ResNet-18, the spiking neurons after the addition operation in each residual block are moved in front of the addition operation, which is more beneficial for the spiking operation \cite{MLF,SEWResNet}.
\begin{table*}[!h]
 \centering
 \begin{tabular}{ccc}
  \toprule
  Stage & VGG-9 & ResNet-18 \\
  \midrule
  1 & - & Conv($3 \times 3@64$)\\
  \hline
  1  &  \makecell{Conv($3 \times 3$@64) \\ Conv($3 \times 3$@128)} &
  \makecell{
  $\left(
 	    \begin{array}{cc}  
 			 \makecell{Conv(3 \times 3@64) \\Conv(3 \times 3@64)}
        \end{array}
    \right)\times 2
  $}\\
  \hline
    & average pool(stride=2) & -\\
  \hline
  2  & \makecell{Conv($3 \times 3$@256) \\ Conv($3 \times 3$@256)} & 
  $\left(
 	    \begin{array}{cc}  
 			 \makecell{Conv(3 \times 3@128) \\ Conv(3 \times 3@128)}
        \end{array}
    \right)\times 2
  $\\
  \hline
    & average pool(stride=2) & -\\
  \hline
  3  & \makecell{Conv($3 \times 3$@512) \\ Conv($3 \times 3$@512)} & $\left(
 	\begin{array}{cc}  
 			 \makecell{Conv(3 \times 3@256) \\ Conv(3 \times 3@256)}
 \end{array}
 \right)\times 2$\\
  \hline
    & average pool(stride=2) & -\\
  \hline
  4  & \makecell{Conv($3 \times 3$@512) \\ Conv($3 \times 3$@512)} & $\left(
 	\begin{array}{cc}  
 			 \makecell{Conv(3 \times 3@512) \\ Conv(3 \times 3@512)}
 \end{array}
 \right)\times 2$\\
  \hline
    \multicolumn{3}{c}{global average pool, fc}\\
  \bottomrule
 \end{tabular}
  \caption{Structures of VGG-9 and ResNet-18, where fc denotes the fully connected layer.}
 \label{model}
\end{table*}

\subsection{B Accuracy Change Curves}
\begin{figure*}[!t]
\centering
\subfigure[VGG-9]
	{
	\includegraphics[width=0.31\textwidth]{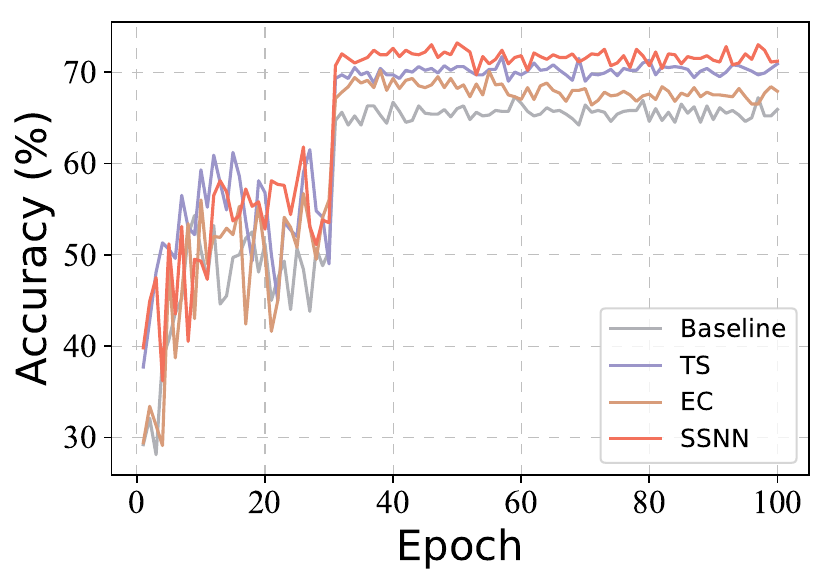}
	\includegraphics[width=0.31\textwidth]{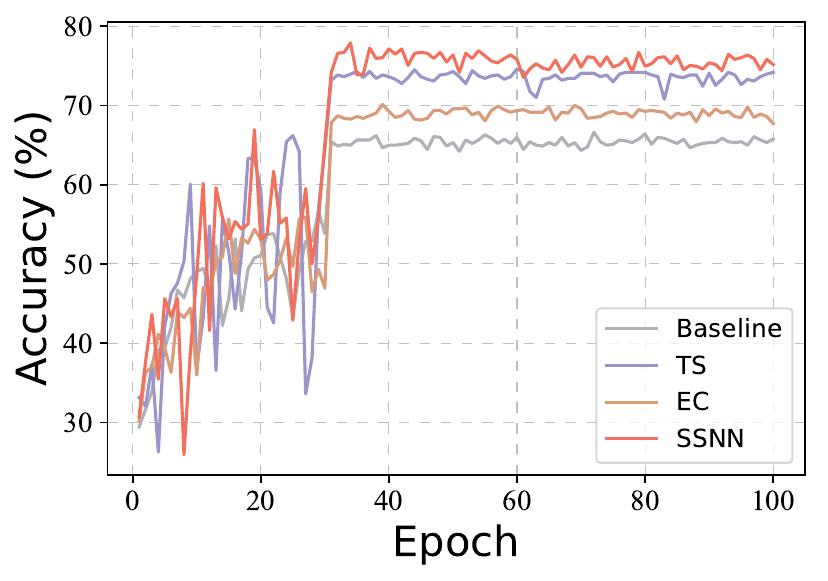}
	\includegraphics[width=0.31\textwidth]{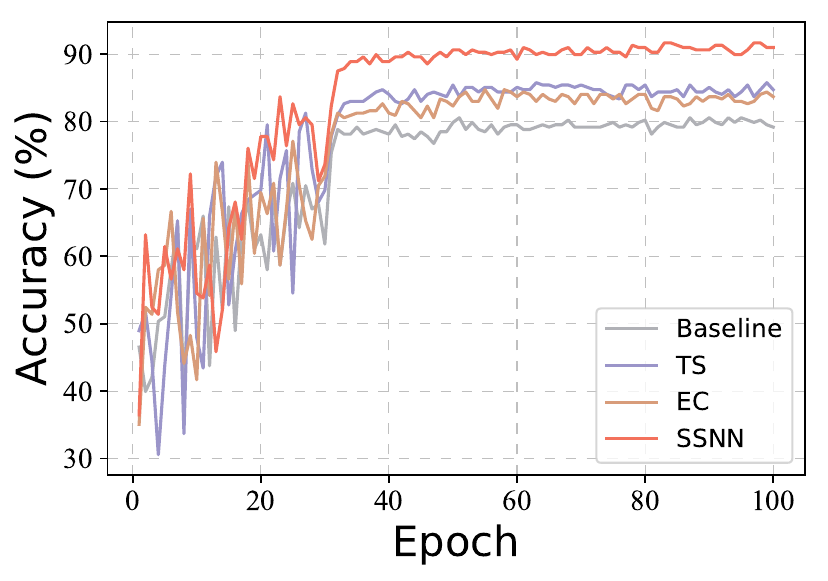}
	}
	\subfigure[ResNet-18]
	{
	\includegraphics[width=0.31\textwidth]{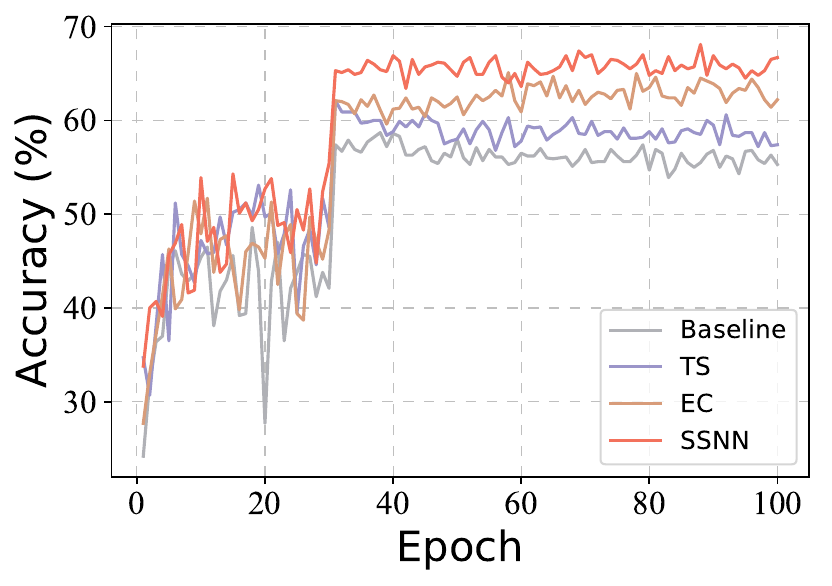}
	\includegraphics[width=0.31\textwidth]{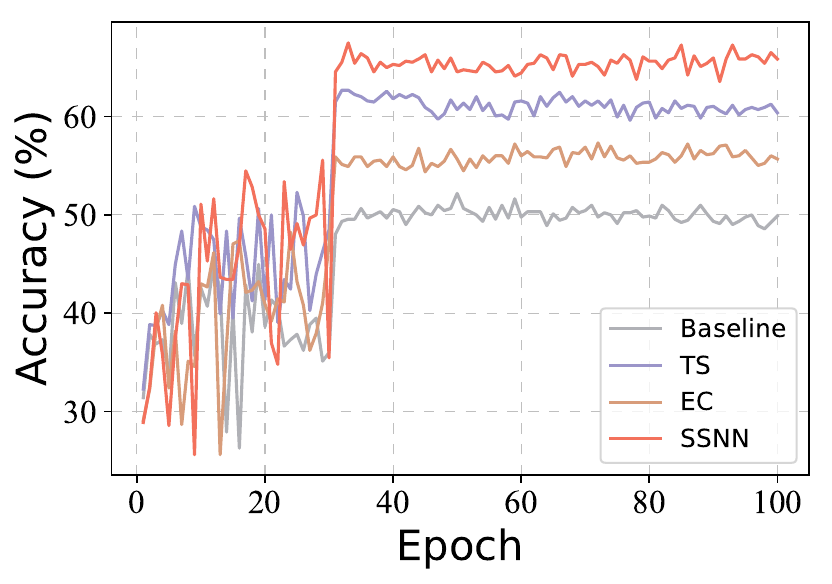}
	\includegraphics[width=0.31\textwidth]{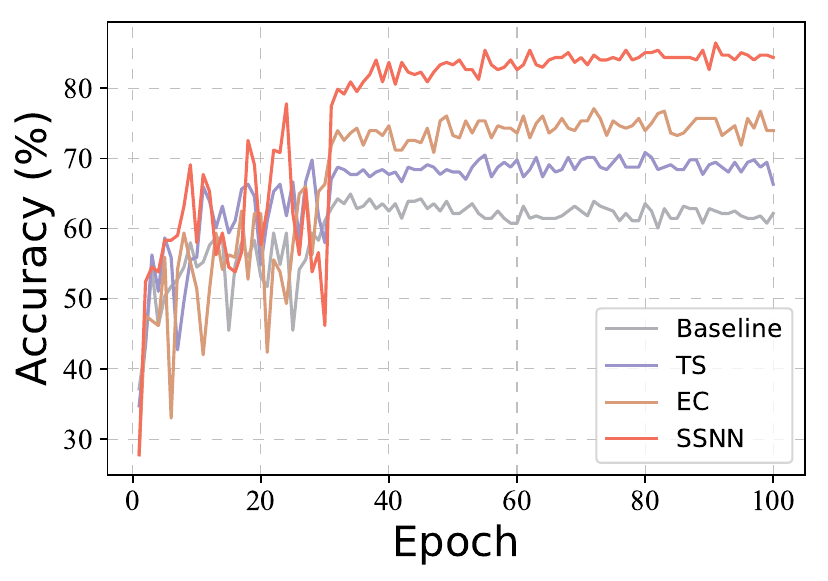}
	}
\caption{Accuracy curves during training (left: CIFAR10-DVS, middle: N-Caltech101, right: DVS-Gesture). }
\label{accuracy}
\end{figure*}
The accuracy curves of the training process are shown in Fig. \ref{accuracy}, where the differences between the proposed methods and the baseline can be perceived more clearly. In Fig. \ref{accuracy}, the convergence accuracies of the proposed methods are consistently above the baseline, confirming the superior performance of the proposed methods, a conclusion consistent with Table 2 in the Experiments section.

\subsection{C Details of Reproduction of Existing Methods}
For reproducing the existing methods, the network structure, hyperparameters, and training method are the same as our SSNN, if not otherwise specified.

\textbf{PLIF:} We set the membrane potential time constant $\tau$ of LIF neurons in PLIF \cite{PLIF} as a learnable parameter, with each layer of neurons having the same $\tau$. Training starts with an initial $\tau$ value of 2.0.

\textbf{tdBN:} We replaced the Batch Normalization layer in the VGG-9 network with tdBN \cite{tdBN_2021}, leaving the rest of the parameters and structure unchanged.

\textbf{BackEISNN:} We replaced the LIF neurons in the VGG-9 network with LIF neurons with adaptive self-feedback mechanism and excitatory/inhibitory balance to achieve the BackEISNN \cite{BackEISNN}, leaving the rest of the parameters and structure remained unchanged. Both the adaptive self-feedback mechanism and the excitatory/inhibitory balance mechanism were implemented using the $3 \times 3$ convolution.

\textbf{MLF:} We replaced the LIF neurons in the VGG-9 network with three levels of MLF \cite{MLF} neurons with firing thresholds of 0.6, 1.6, and 2.6, respectively, and kept the rest of the parameters and structure unchanged.

\textbf{TEBN:} We replaced the Batch Normalization layer in the VGG-9 network with TEBN \cite{TEBN}, leaving the rest of the parameters and structure unchanged.

\textbf{Spikformer:} In reproducing Spikformer \cite{Spikformer}, we use the same approach as in the original paper, i.e., data augmentation of the CIFAR10-DVS, using data of $128 \times 128$ size without downsampling directly into Spikformer. The loss function, learning rate adjustment policy, and batch size employed in the original paper are used. For N-Caltech101 and DVS-Gesture, the data augmentation is not used. In particular, the data of N-Caltech101 is resized to $128 \times 128$. Training a total of 106 epochs on the three datasets with Spikformer was the same configuration as the original paper on  CIFAR10-DVS. During training, Spikformer works with the same low latency as our SSNN (timestep of 5).

\subsection{D Stage-Wise Timestep Setting}
For VGG-9, we explored the performance of SSNN at different average timesteps, and the specific stage-wise timesteps are shown in Table \ref{timestep setting}. Note that for each fixed average timestep in Table \ref{timestep setting}, one setting is given as an example, and several different settings are actually available.
\begin{table}[!tb]
 \centering
 \begin{tabular}{ccccccccc}
 \hline
 Average timestep & 3 & 4 & 5 & 6 & 7 & 8 & 9 & 10\\
  \hline
  $T_1$ & 5 & 7 & 8 & 10 & 12 & 12 & 14 & 16\\
  $T_2$ & 4 & 5 & 6 & 7 & 8 & 9 & 10 & 12\\
  $T_3$ & 2 & 3 & 4 & 5 & 6 & 7 & 8 & 8\\
  $T_4$ & 1 & 1 & 2 & 2 & 2 & 4 & 4 & 4\\
  \hline
 \end{tabular}
\caption{Stage-wise timestep setting at average timestep range from 3 to 10.}
 \label{timestep setting}
\end{table}

\subsection{E Mutual Enhancement with Spikformer}
In addition, we investigate the performance of our method at relatively high timesteps and whether it can be mutually enhanced with Spikeformer. We evaluate the performance of our method on DVS-Gesture at 16 timesteps. We divide the Spikformer into two stages based on the Spikformer encoder block and set the timestep of the first stage equal to that of the original Spikformer to achieve a fair comparison (at which point our method has a lower overall latency). The experimental results are shown in Table~\ref{highlatency}.

\begin{table}[h]
\tabcolsep=0.004\columnwidth
 \centering
 \begin{tabular}{ccc}
  \toprule
Spikformer(16) & VGG-9(10)\dag &Spikformer(10)\dag\\
95.25\% & 95.26\%  &95.53\% \\
  \hline
Spikformer(16)* & Spikformer(10)*\dag & Spikformer(16)*\dag\\
97.42\%  & 96.18\%  & 98.03\% \\
  \bottomrule
 \end{tabular}
 \caption{Average test accuracy on DVS-Gesture. The average timestep is shown in brackets. }
 \label{highlatency}
\end{table}
When our method (denoted by \dag) is integrated with VGG-9 and Spikformer , its stage-wise timesteps are \{16,12,8,4\} and \{16,4\}, respectively, and both achieve higher average accuracies than the original Spikformer. Our data were processed in 30ms integration, and for a fairer comparison we used the same data pre-processing method as Spikformer paper (averaging integration), denoted by *. At this point, the performance of our Spikformer(10)*\dag is slightly lower than the original Spikformer(16)*, but reaches 98.03\% accuracy when the timestep is increased to \{20,12\} (Spikformer(16)*\dag), which improves the performance of the original Spikformer. This further demonstrates the generalizability of our method to a wide range of network architectures.

\bibliography{SSNN}
\end{document}